\documentclass[runningheads]{llncs}
\usepackage{graphicx}
\usepackage{comment}
\usepackage{amsmath,amssymb} 
\usepackage{color}
\usepackage{multirow}

\begin{document}
	\pagestyle{headings}
	\mainmatter
	\def\ECCVSubNumber{4010}  
	
	\title{Challenge-Aware RGBT Tracking} 

	\titlerunning{Challenge-Aware RGBT Tracking}
	%
	\author{Chenglong Li\orcidID{0000-0002-7233-2739} \and
	Lei Liu\orcidID{0000-0003-2749-5528} \and
	Andong Lu\orcidID{0000-0002-0902-2260} \and
   Qing Ji\orcidID{0000-0002-2466-9193} \and
   Jin Tang\orcidID{0000-0002-4123-268X}}
	\authorrunning{C. Li et al.}
	%
	\institute{Key Lab of Intelligent Computing and Signal Processing of Ministry of Education \\
	Anhui Provincial Key Laboratory of Multimodal Cognitive Computation \\
School of Computer Science and Technology, Anhui University, Hefei 230601, China
	\email{\{lcl1314, adlu\_ah\}@foxmail.com, \{liulei970507, m18815684602\}@163.com, tangjin@ahu.edu.cn}}
	\maketitle
	
	\begin{abstract}
		RGB and thermal source data suffer from both shared and specific challenges, and how to explore and exploit them plays a critical role to represent the target appearance in RGBT tracking.
		In this paper, we propose a novel challenge-aware neural network to handle the modality-shared challenges (e.g., fast motion, scale variation and occlusion) and the modality-specific ones (e.g., illumination variation and thermal crossover) for RGBT tracking.
		In particular, we design several parameter-shared branches in each layer to model the target appearance under the modality-shared challenges, and several parameter-independent branches under the modality-specific ones.
		Based on the observation that the modality-specific cues of different modalities usually contains the complementary advantages, we propose a guidance module to transfer discriminative features from one modality to another one, which could enhance the discriminative ability of some weak modality.
		Moreover, all branches are aggregated together in an adaptive manner and parallel embedded in the backbone network to efficiently form more discriminative target representations.
		These challenge-aware branches are able to model the target appearance under certain challenges so that the target representations can be learnt by a few parameters even in the situation of insufficient training data.
		From the experimental results we will show that our method operates at a real-time speed while performing well against the state-of-the-art methods on three benchmark datasets.
		
	\end{abstract}

	\section{Introduction}
	
	The task of RGBT tracking is to deploy complementary benefits of RGB and thermal infrared information to estimate the states (i.e., location and size) of a specified target in subsequent frames of a video sequence given the initial state in the first frame.
	Recently, it becomes increasingly popular due to its potential value in all-day all-weather applications such as surveillance and unmanned driving.
	Although RGBT tracking has achieved many breakthroughs, it still remains unsolved partly due to various challenges including illumination variation, thermal crossover and occlusion, to name a few.
	
	\begin{figure}[t]
		\centering
		\includegraphics[width=0.5\textwidth]{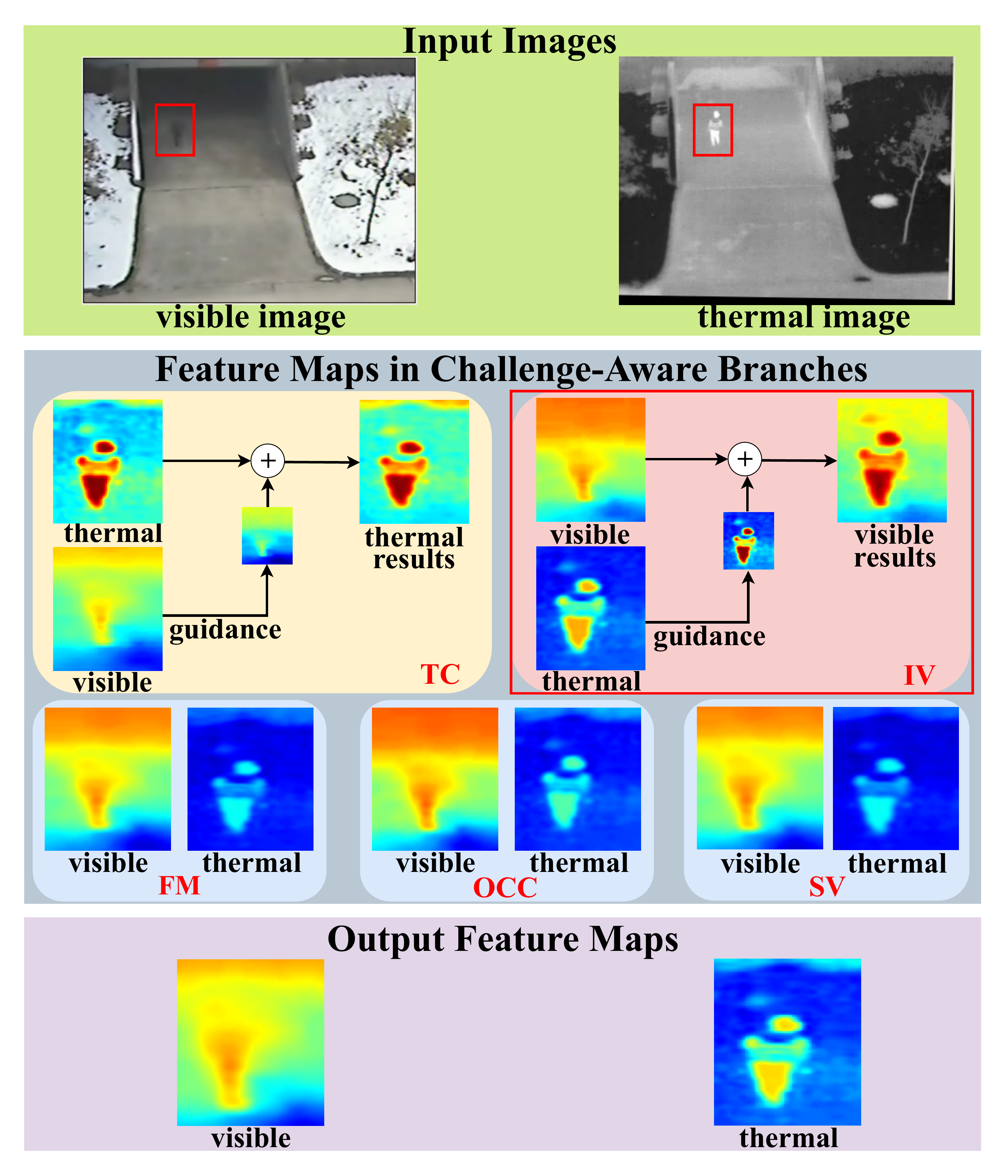} 
		\caption{Illustration of our challenge-aware RGBT tracker in handling an example frame pair with the illumination variation (IV). 
			Herein, thermal crossover (TC) and illumination variation (IV) are the modality-specific challenges, and we use the guidance module to enhance target representations of visible modality through the guidance of thermal modality (see the IV block for details).
			Meanwhile, a gate scheme is used to avoid noisy information in the guidance module (see TC block for details).
			Fast motion (FM), occlusion (OCC) and scale variation (SV) are the modality-shared challenges.
			The output feature maps are obtained by adaptively aggregating all challenge-aware target representations. }
		\label{fig:motivation}
	\end{figure}
	
	Numerous effective algorithms have been proposed to address the problem of RGBT tracking, from simple weighting fusion and sparse representation to deep learning techniques.
	At present, deep learning trackers have dominated this research field.
	These trackers could be categorized into three classes, including multimodal representation models (e.g., MANet~\cite{long2019multi}), multimodal fusion models (e.g., mfDiMP~\cite{zhang2019multi}) and hybrid of them (e.g., DAPNet~\cite{zhu2019dense}).
	Although these algorithms have achieved great success in RGBT tracking, but not take into account the target appearance changes under different challenges which might limit tracking performance. 

	To handle these problems, we propose a novel Challenge-Aware RGBT Tracker (CAT) which exploits the annotations of challenges to learn robust target representations under different challenges even in the case of insufficient training data.
   Qi et al.~\cite{qi2019learning} design an interesting CNN model that embeds attribute-based representations for effective single-modality tracking.
	Different from it, existing RGBT tracking datasets include five major challenges annotated manually for each video frame, including illumination variation (IV), fast motion (FM), scale variation (SV), occlusion (OCC) and thermal crossover (TC).
	We find that some of them are modality-shared including FM, SV and OCC, and remaining ones are modality-specific including IV and TC.
	To better deploy these properties, we propose two kinds of network structures.
	For the modality-shared challenges, the target appearance under each one is modeled by a convolutional branch across all modalities.
	For the modality-specific challenges, the target appearance under each one is modeled by a convolutional branch in each modality.
	The modality-specific branches of different modalities usually contains the complementary advantages in representing the targets.
	Therefore, we design a guidance module to transfer discriminative features from one modality to another one.
	In particular, we design a gated point-wise transform layer which enhances the discriminative ability of some weak modality while avoiding the propagation of noisy information.

	All challenge-aware branches are aggregated together in an adaptive manner and parallel embedded in the backbone network to efficiently form more discriminative target representations.
	These branches are able to model the target appearance under certain challenge in the form of residual information and only a few parameters are required in learning target appearance representations. 
	The issue of the failure to capture target appearance changes under different challenges with less training data in RGBT tracking is therefore addressed.
	The effectiveness of our challenge-aware tracker is shown in Fig.~\ref{fig:motivation}.

	In the training phase, there are three problems to be considered.
	First, the classification loss of a training sample with any attribute will be backwardly propagated to all challenge branches.
	Second, the training of the modality-specific branches should not be the same with the modality-shared ones as they contain additional guidance modules.
	Third, the challenge annotations are available in training stage but unavailable in test stage.
	To handle these problems, we propose a three-stage scheme to effectively train the proposed network.
	In the first stage, we remove all guidance modules as well as adaptive aggregation layers and train all challenge-aware branches one-by-one.
	In the second stage, we remove all adaptive aggregation layers and only train all guidance modules in the modality-specific challenge branches.
	In the third stage, we use all challenging and non-challenging frames in training dataset to learn the adaptive aggregation layers and classifier, and fine-tune the parameters of backbone network at the same time.
	Experimental results show that the proposed three-stage training scheme is effective.
	
	We summarize the major contributions of this paper as the following aspects. 
	First, we propose an effective deep learning framework based on a novel challenge-aware neural network to handle the problem of the failure to fully model target appearance changes under different challenges even in the case of insufficient training data in RGBT tracking.
	Second, we propose two kinds of network structures to model the target appearance under the modality-shared challenges and the modality-specific challenges respectively for learning robust target representations even in the presence of some weak modality.
	Third, with both efficiency and effectiveness considerations, we design the parallel and hierarchical architectures of the challenge-aware branches and embed them in the backbone network in the form of residual information, which can be learned with a few parameters.
	Finally, extensive experiments on three benchmark datasets show that our tracker achieves the promising performance in terms of both efficiency and effectiveness against the state-of-the-art methods. 

	\section{Related Work}
	
	\subsection{RGBT Tracking Methods}
	Deep learning trackers have dominated the research field of RGBT tracking.
	Li et al.~\cite{li2018fusing} is the first to apply deep learning technique to RGBT tracking, and propose a two-stream CNN and a fusion subnetwork to extract features of different modalities and perform adaptive fusion respectively. 
	To better fuse features of RGB and thermal data, Zhu et al.~\cite{zhu2019dense} propose a network to aggregate features of all layers and all modalities, and then prune these features to reduce noises and redundancies.
	Gao et al.~\cite{gao2019deep} incorporate attention mechanisms in the fusion to suppress noise of modalities.
	To further improve the capability of RGBT feature representation, Li et al.~\cite{long2019multi} propose a multi-adapter architecture for learning modality-shared, modality-specific and instance-aware target representations respectively.
	Zhang et al.~\cite{zhang2019multi} use different levels of fusion strategies in an end-to-end deep learning framework and achieve promising tracking performance.

	\subsection{Multi-Task Learning}
	Multi-task learning in computer vision aims to solve multiple visual tasks in a single model. 
	A typical setting is to share early layers of the network and then build multiple branches at the last layer for implementing different tasks. 
	It is shown in ANT~\cite{qi2019learning} that shares weights in early layers and builds multi-branches after last layer for learning different attribute representations. 
	Rebuffi et al.~\cite{rebuffi2017learning} propose a series residual adapters module to build the networks with a high-degree of parameter sharing for multi-task learning. 
	The residual adapter can deal various visual domains by fine-tuning a small number of parameters in all layer. 
	Then, they improves their work by replacing the serial residual adapter with a parallel structure and achieves better performance in both accuracy and computational complexity~\cite{rebuffi2018efficient}.
	
	\begin{figure}[t]
		\centering
		\includegraphics[width=1\textwidth]{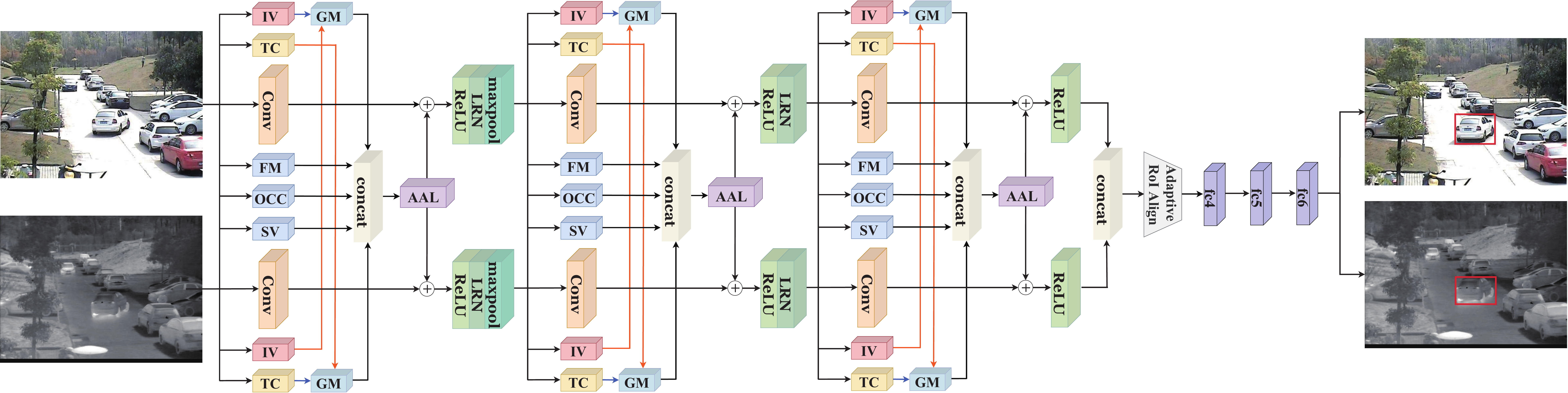} 
		\caption{Network architecture of the proposed challenge-aware RGBT tracker. 
			Herein, $+$  denotes the operation of element-wise addition.
			The abbreviations of GM and AAL denote the guidance module and adaptive aggregation layer respectively.
			The challenge abbreviations of IV, TC, FM, OCC and SV are illumination variation, thermal crossover, fast motion, occlusion and scale variation respectively. }
		\label{fig:network}
	\end{figure}
	%
	\section{Challenge-Aware RGBT Tracker}
	In this section, we first present the details of the proposed challenge-aware neural network, and the respective progressive learning algorithm.
	Then, we describe the online tracking method based on the challenge-aware network.
	\subsection{Challenge-Aware Neural Network}
	\subsubsection{Overview}
	As discussed in previous section, there is failure to learn the target appearance representation under different challenges which limits RGBT tracking performance.
	To handle this problem, we exploit the annotations of challenges in existing RGBT tracking datasets and propose multiple challenge-aware branches to model the target appearance under certain challenges. 
	To account for the properties of different challenges in RGBT tracking, all challenges are separated into the modality-specific ones and modality-shared ones, and we propose two kinds of network structures to model them respectively.
	Moreover, we design an adaptive aggregation module to adaptively combine all challenge-aware representations even without knowing challenges for each frame in tracking process, and can also handle situations with multiple challenges in one frame.
	%
	%
	In order to develop CNN ability of multi-level feature expression, we add challenge-aware branches into each layers of the backbone network with hierarchical architecture.
	In summary, our challenge-aware neural network consists of five components, including two-stream CNN backbone, modality-shared challenge branches, modality-specific challenge branches, adaptive aggregation module of all branches and hierarchical architecture, as shown in Fig.~\ref{fig:network}.
	We present the details of these components in the following.
	
	\subsubsection{Two-stream CNN backbone}
	As other trackers adopted, we select a lightweight CNN to extract target features of two modalities for the tracking task.
	In specific, we use a two-stream CNN to extract RGB and thermal representations in parallel, and each composes of three convolutional layers modified from the VGG-M~\cite{chatfield2014return}.
	Herein, the kernel sizes of three convolutional layers are $7\times 7$, $5\times 5$ and $3\times 3$ respectively.
	The max pooling layer in the second block is removed and the dilate convolution~\cite{yu2015multi} is introduced in last convolutional layer with the dilate ratio as $3$ to enlarge the resolution of output feature maps.
	To improve the efficiency, we introduce the RoIAlign pooling layer to allow features of candidate regions be directly extracted on feature maps, which greatly accelerates feature extraction~\cite{jung2018real} in tracking process.
	After that, three fully connected layers (fc4-6) are used to accommodate appearance changes of instances in different videos and frames.
	Finally, we use the softmax cross-entropy loss and instance embedding loss~\cite{jung2018real} to perform the binary classification to distinguish the foreground and background. 
	
	\begin{figure}[t]
		\centering
		\includegraphics[width=0.8\textwidth]{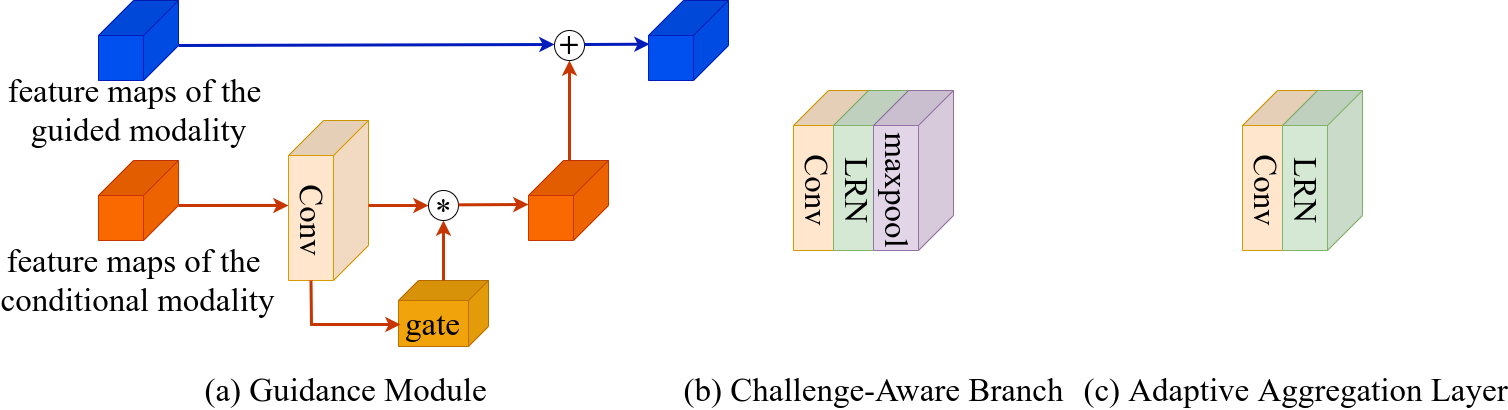} 
		\caption{Structures of three subnetworks in our challenge-aware neural network.}
		\label{fig:sub_network}
	\end{figure}
	
	\subsubsection{Modality-shared branches}
	Existing RGBT tracking datasets include five major challenges annotated manually for each video frame, including illumination variation (IV), fast motion (FM), scale variation (SV), occlusion (OCC) and thermal crossover (TC). 
	Note that more challenges could be considered in our framework, and we only consider the above ones and the tracking performance is improved clearly as shown in the experiments.
	We find that some of them are modality-shared including FM, SV and OCC, and remaining ones are modality-specific including IV and TC. 
	To better deploy these properties, we propose two kinds of network structures.
	We first describe the details of the network structure for the modality-shared challenges.
	For one modality-shared challenge, the target appearance can be modeled by a same set of parameters to capture the collaborative information in different modalities.
	To this end, we design a parameter-shared convolution layer to learn the target representations under a certain modality-shared challenge.
	To reduce the number of parameters of modality-shared branches, we design a parallel structure that adds a block with small convolution kernels on the backbone network, as shown in Fig.~\ref{fig:network}.
	Although only small convolution kernels are used, such design is able to encode the target information under modality-shared challenges effectively. 
	Since different modality-shared branches should share a larger portion of their parameters, the number of modality-shared parameters should be much smaller than the backbone.
	In specific, We use two convolution layers with the kernel size of $3\times 3$ to represent the challenge-aware branches in first convolution layer, and one convolution layer with the kernel size of $3\times 3$ and $1\times 1$ in second and third layers respectively. 
	For all modality-shared branches, the Local Response Normalization (LRN) is used after convolution operation to accelerate the speed of convergence and improve the generalization ability of the network.
	In addition, the operation of max pooling is used to make the resolution of feature maps obtained by the modality-shared branches the same with that extracted by the corresponding convolution layer in the backbone network. 
	Fig.~\ref{fig:sub_network} (b) shows the details of the modality-shared branch.

	\begin{figure}[t]
		\centering
		\includegraphics[width=0.8\textwidth]{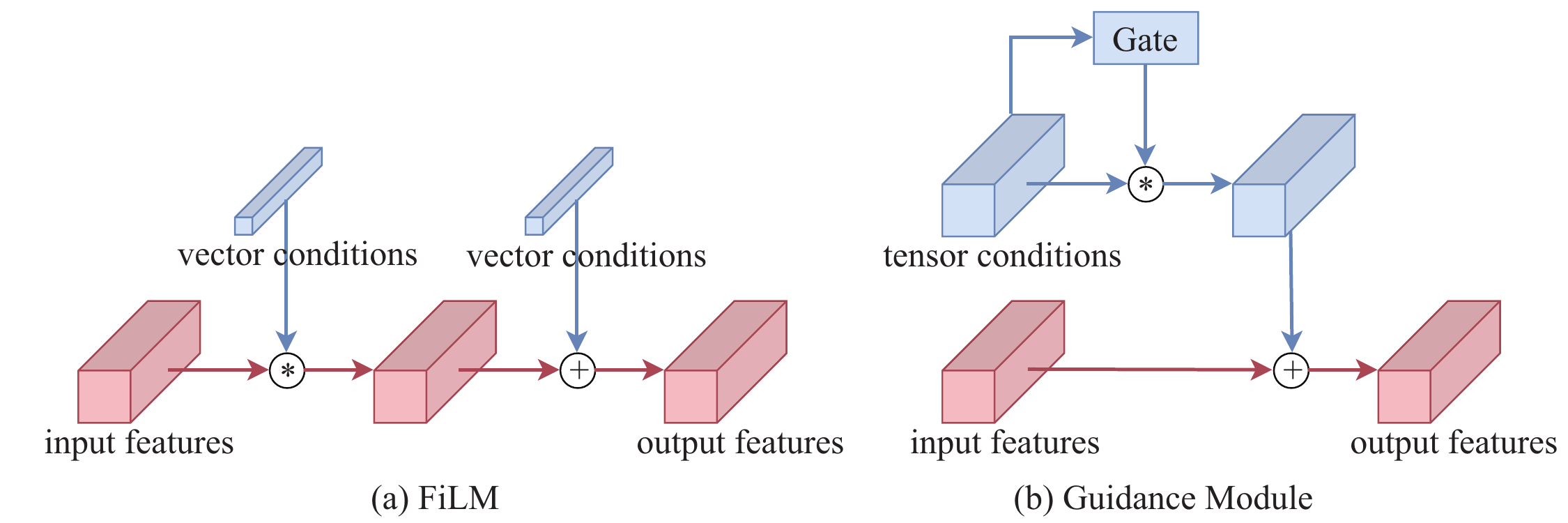} 
		\caption{Differences of our guidance module from FiLM~\cite{perez2018film}. 
			Herein, $+$ and $*$ denote the operation of element-wise addition and multiplication respectively.
			We can see that our guidance module only use the feature shift to perform guiding as our task is simpler than~\cite{perez2018film}, and the experiments justify the effectiveness of our such design.
			In addition, we introduce the gate scheme and point-wise linear transformation in our task while FiLM is the channel-wise linear transformation.}
		\label{fig:guidance}
	\end{figure}

	\subsubsection{Modality-specific branches}
	As discussed above, the modality-shared branches are used to model the target appearance under one challenge across all modalities for the collaboration.
	To take the heterogeneity into account, we propose modality-specific branch to model the target appearance under one challenge for each modality.
	The structure of the modality-specific branch is the same with the modality-shared one, as shown in Fig.~\ref{fig:sub_network} (b).
	Different from the modality-shared branches, the modality-specific ones usually contains the complementary advantages of different modalities in representing the target, and how to fuse them plays a critical role in performance boosting.
	For example, in IV, the RGB data is usually weaker than the thermal data.
	If we improve the target representations in the RGB modality using the guidance of the thermal source, the tracking results would be improved as the target features are enhanced.
	To this end,  we design a guidance  module  to  transfer discriminative  features  from  one  modality  to  another  one.
	
	The structure of the guidance module is shown in Fig.~\ref{fig:sub_network} (a).
	Our design is motivated by FiLM~\cite{perez2018film} that introduces the feature-wise linear modulation to learn a better feature maps with the help of condition information in the task of visual reasoning.
	It is implemented by a Hadamard product with priori knowledge and adding a conditional bias which play a roles of feature-wise scale and shift respectively.
	Unlike processing text and visual information in FiLM, our goal is simpler and only needs to improve the discrimination of features in some weak modality from help of another one. 
	Moreover, for some visual tasks like object tracking, the spatial information is crucial for accuracy location and thus should be considered in feature modulation~\cite{DSFT18cvpr}.
	Taking these into considerations, we use a point-wise feature shift to transfer discriminative information from one modality to another, and the differences of our guidance module from FiLM could be found in Fig.~\ref{fig:guidance}.
	Moreover, we introduce a gate mechanism to suppress the spread of noise information in the feature propagation which can be verified by Fig.~\ref{fig:motivation} in the case of TC.
	In the design, a convolution layer with the kernel size of $1\times 1$  followed by a nonlinear activation layer are used to learn a nonlinear mapping, and the gate operation is implemented by element-wise sigmoid activation, as shown in Fig.~\ref{fig:sub_network} (a). 
	The formulation of our guidance module is as follows:
	\begin{equation}
	\begin{aligned}
	&\gamma = w_1\ast {\bf x}+b_1,\\
	&\beta = w_2\ast ReLU(\gamma)+b_2,\\
	&\tilde{\beta} = \sigma(\beta)\ast \gamma,\\
	&{\bf z} = {\bf z}+\tilde{\beta}\\
	\end{aligned}
	\label{eq::guidance}
	\end{equation}
	where $w_i$ and $b_i (i=1, 2)$ represent the weight and bias of the convolutional layer respectively.
	${\bf x}$ and ${\bf z}$ denote the feature maps of the prior and guided modalities respectively, and $\sigma$ is the sigmoid function.
	$\gamma$ and $\tilde{\beta}$ denote the point-wise feature shift without and with the gate operation respectively.
	
	\subsubsection{Adaptive aggregation module}
	Since it is unknown what challenges each frame has in tracking process, we need to design an adaptive aggregation module to combine all branches effectively and form more robust target representations, and the structure is shown in Fig.~\ref{fig:sub_network} (c). 
	In the design, we use the concatenate operation rather than the addition to aggregate all branches to avoid the dispersion of differences in these branches in the adaptive aggregation layer. 
	Then, the convolution layer with kernel size as $1\times 1$ is used to extract adaptive features and achieve dimension reduction.

	\begin{figure}[t]
		\centering
		\includegraphics[width=0.5\textwidth]{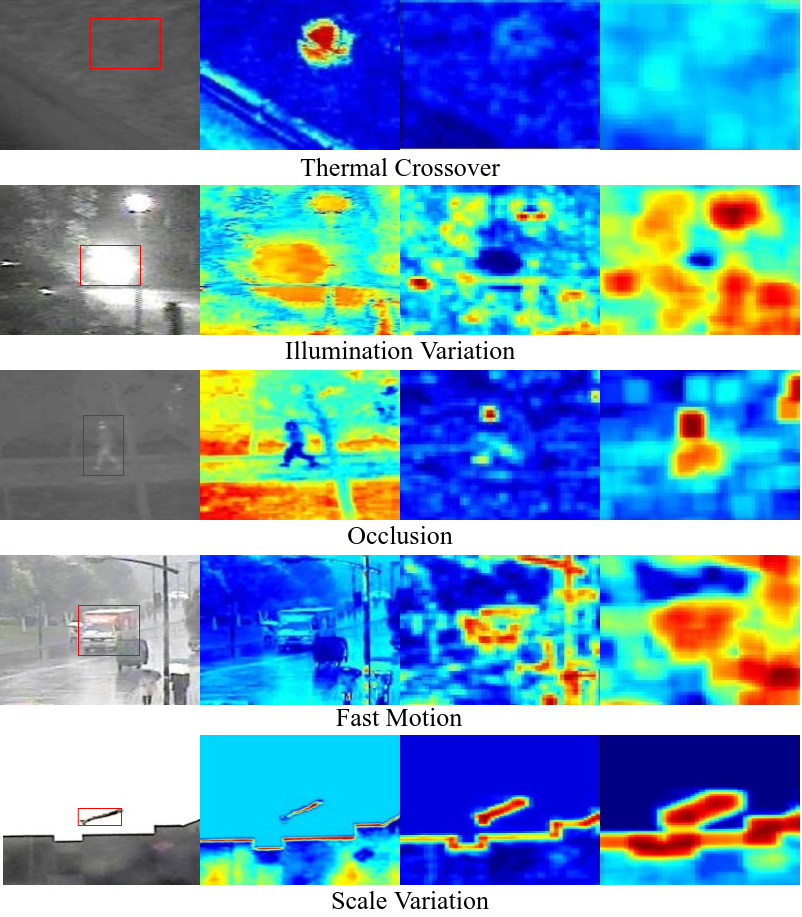} 
		\caption{Illustration of feature maps in different layers and different challenges. 
			We can see that different challenge attributes could be well represented in different feature layers in some scenarios. }
		\label{fig:feature_maps}
	\end{figure}

	\subsubsection{Hierarchical challenge-aware architecture}
	We observe that the target appearance under different challenges could be well represented in different layers, as shown in Fig.~\ref{fig:feature_maps}.
	For example, in some scenarios, the target appearance under the challenge of thermal crossover can be well represented in shallow layers of CNNs, occlusion in middle layers and fast motion in deep layers.
	To this end, we add the challenge-aware branches into each convolutional layer of the backbone network and thus deliver a hierarchical challenge-aware network architecture, as shown in Fig.~\ref{fig:network}.
	Note that these challenge-aware branches are able to model the target appearance under certain challenge in the form of residual information and only a few parameters are required in learning target appearance representations. 
	The issue of the failure to capture target appearance changes under different challenges with less training data in RGBT tracking is therefore addressed.
	
	\subsection{Training Algorithm}
	In the training phase, there are three problems to be addressed. 
	First, the classification loss of a training sample with any attribute will be backwardly propagated to all challenge branches. 
	Second, the training of the modality-specific branches should not be the same with the modality-shared ones as they contain additional guidance modules. 
	Third, the challenge annotations are available in training stage but unavailable in test stage. 
	Therefore, we propose a three-stage training algorithm to effectively train the proposed network.
	
	{\flushleft \bf Stage I: Train all challenge-aware branches.}
	In this stage, we remove all guidance modules and adaptive aggregation modules, and train all challenge-aware branches (including modality-shared and modality-specific) using the challenge-based training data.
	In specific, we first initialize the parameters of our two-stream CNN backbone by the pre-trained model in VGG-M~\cite{chatfield2014return}, and these parameters are fixed in this stage. 
	The parameters of all challenge-aware branches and fully connection layers are randomly initialized and the learning rates are set to 0.001 and 0.0005 respectively. 
	The optimization strategy we adopted is the stochastic gradient descent (SGD) method with the momentum as 0.9, and we set the weight decay to 0.0005.
	The number of training epochs is set to 1000. 
	
	{\flushleft \bf Stage II: Train all guidance modules.}
	After each challenge branches are trained in stage I,  it is necessary for modality-specific challenge branches to learn the guidance module separately to solve the problem of weak modality. All of hypr-parameters are set the same as stage I. 
	{\flushleft \bf Stage III: Train all adaptive aggregation modules.}
	In this stage, we use all challenging and non-challenging frames to learn the adaptive aggregation modules and classifier, and fine-tune the parameters of backbone network at the same time. 
	To be specific, we fix the parameters of all challenge branches and guidance modules pre-trained in first two stages.
	The learning rates of adaptive aggregation modules and fully connection layers are set to 0.0005, and set to 0.0001 in backbone network.
	We adopt the same optimization strategy with Stage I, and the number of epochs is set to 1000.

	\subsection{Online Tracking}
	In the first frame with the initial bounding box, we collect 500 positive samples and 5000 negative samples, whose IoUs with the initial bounding box are greater than 0.7 and less than 0.3 respectively.
	We use these samples to fine-tune the parameters of the fc layers in our network to adapt to the new tracking sequence by 50 epochs, where the learning rate of the last fc layer (fc6) is set to 0.001 and others (fc4-5) are 0.0005. 
	In addition, 1000 bounding boxes whose IoUs with the initial bounding box are larger than 0.6 are extracted to train the bounding box regressor and the hyper-parameters are the same as the above.
	Starting from the second frame, if the tracking score is greater than a predefined threshold (set to 0 empirically), we think the tracking is success.
	In this case, we collect 20 positive bounding boxes whose IoUs with present tracking result are larger than 0.7 and 100 negative samples whose IoUs with present tracking result are less than 0.3 for online update to adapt to appearance changes of the target during tracking process. 
	The long-term update is conducted every 10 frames, the learning rate of the last fc layer (fc6) is set to 0.003 and others (fc4-5) are 0.0015 and the number of epochs are set to 15. 
	And the short-term update is conducted when tracking is failed in current frame and the hyper-parameters of training are the same with in the long-term update~\cite{jung2018real}.
	
	When tracking the $t$-th frame, 256 candidate regions are sampled by Gaussian distribution around the tracking results of the $t-1$-th frame, and then we use the trained network to calculate the scores of these candidate regions, which can be divided into positive samples and negative samples. 
	The candidate region sample with the highest positive score is selected as the tracking result of $t$-th frame.
	In addition, the bounding box regression method is used to fine-tune the tracking results to locate the targets more accurately.
	More details can be referred to MDNet~\cite{nam2016learning}.

	\section{Performance Evaluation}
	In this section, we evaluate our CAT on three benchmark datasets comparing with some state-of-the-art trackers, 
	and the contents contain experimental setting, quantitative comparison and analysis on three RGBT tracking benchmark datasets, 
	and in-depth analysis of our CAT including ablation study and runtime analysis.
	
	\subsection{Experimental Setting}
	
	{\flushleft \bf Evaluation data}.
	We evaluate our CAT on three RGBT tracking benchmark datasets, i.e., GTOT~\cite{li16gtot}, RGBT210~\cite{Li17rgbt210} and RGBT234~\cite{li19rgbt234}. 
	{\bf GTOT:} It is the first standard dataset for RGBT tracking, and includes 50 RGBT video sequences with about a total of 15K frames that captured under different scenes and conditions. 
	{\bf RGBT210:} It is larger dataset for RGBT tracking and contains 210 RGBT video pairs with about 210K frames in total, in which 12 attributes are annotated to facilitate analyzing the attribute-based performance.
	{\bf RGBT234:} It is extended from the RGBT210 dataset and is largest dataset for RGBT tracking at present, and contains 234 RGBT video pairs with 234K frames in total. 
	%
	%
	And provides a more accurate annotations and takes into full consideration of various environmental challenges, such as raining, night, cold and hot days. 
	\emph{Note that the dataset in the VOT2019-RGBT tracking challenge is a subset of RGBT234}.

	{\flushleft \bf Evaluation metrics}.
	In the GTOT, RGBT210, the precision rate (PR) and success rate (SR) in the one-pass evaluation (OPE) are adopted as evaluation metrics. 
	It is worth noting that RGBT234 adopts MPR and MSR which use smaller values of PR and SR in different modalities to compute the final PR and SR.

	\begin{figure*}[t]
		\centering
		\includegraphics[width=1\textwidth]{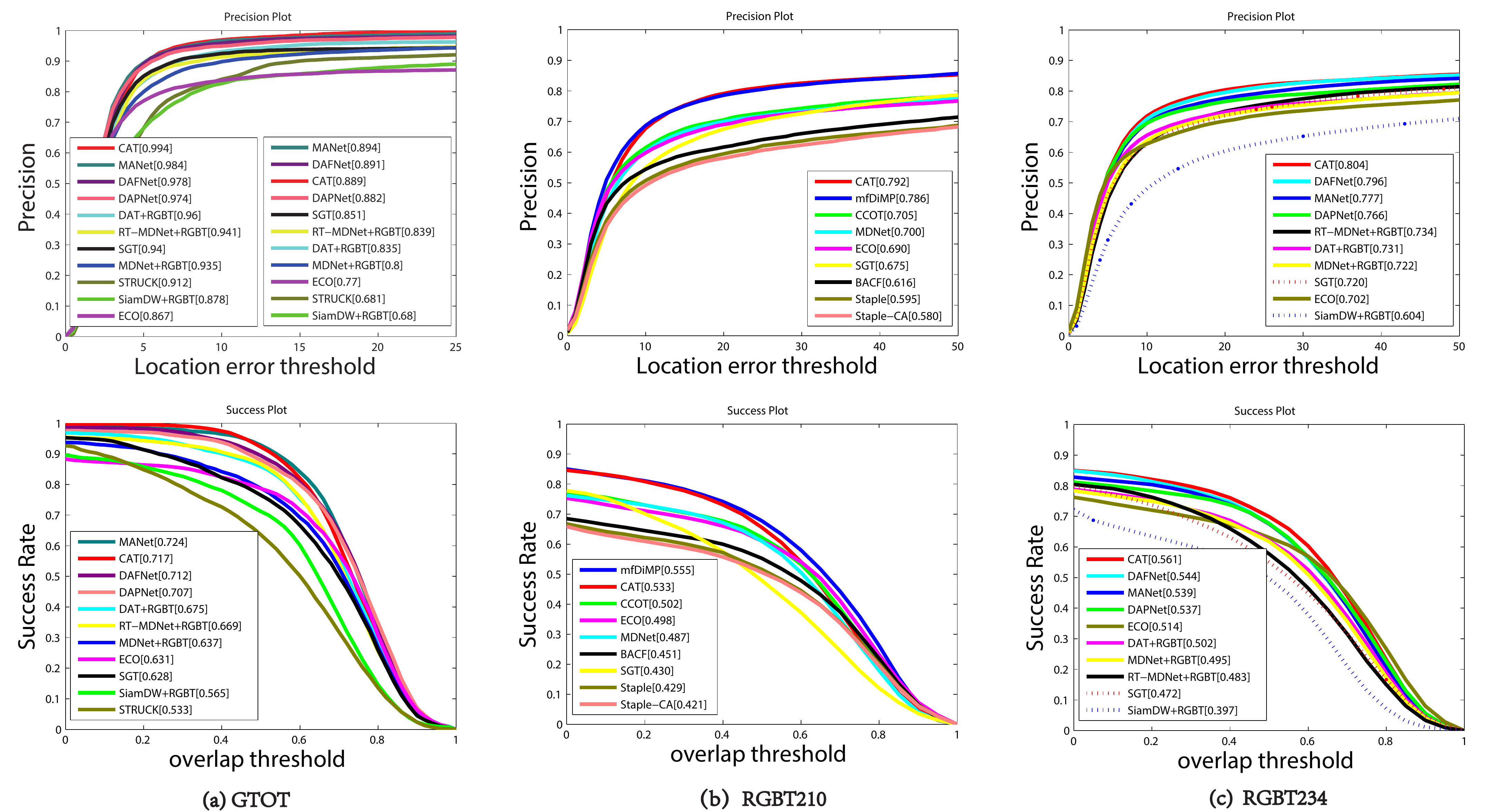} 
		\caption{The evaluation results on GTOT, RGBT210 and RGBT234 datasets. The representative scores of PR/SR is presented in the legend. 
			For the GTOT dataset, we report two representative PR scores with the predefined threshold as 5 pixels (right) and 20 pixels (left) respectively.}
		\label{fig:GTOT+RGBT24+RGBT210}
	\end{figure*}
	
	{\flushleft \bf Implementation details}.
	In the experiments, we test on three benchmark datasets including GTOT, RGBT210 and RGBT234. 
	For the testing on GTOT, we train our challenge-aware branches with the sub-dataset extracted from RGBT234 with annotations of challenge attributes in the stage I and II.
	Then, we use the complete dataset of RGBT234 to train adaptive aggregation layers and classifier. 
	For the testing on RGBT210 and RGBT234, our training dataset is GTOT, and training process is similar to the mentioned above. 
	It is worth noting that in the process of the testing on GTOT, RGBT210 and RGBT234, we set the parameters $grad\_clip$ as 100, 50 and 10 respectively.

	\subsection{Quantitative Comparison}
	To validate the effectiveness of our CAT, we test it on three RGBT benchmark datasets, with some state-of-the-art trackers such as mfDiMP (the winner in VOT2019-RGBT challenge)~\cite{zhang2019multi}, MANet~\cite{long2019multi}, DAFNet~\cite{gao2019deep}, DAPNet~\cite{zhu2019dense}, RT-MDNet+RGBT~\cite{jung2018real} (the baseline tracker we used for implementing CAT), SGT~\cite{Li17rgbt210} and MDNet+RGBT~\cite{nam2016learning}.
	It is worth noting that not all contrast algorithms are RGBT-based tracking, and for the fairness of comparison, the RGB tracking algorithm is extended to RGBT by concatenating the features of the two modalities on the deep layer.
	The overall performance on these three datasets are shown in Fig.~\ref{fig:GTOT+RGBT24+RGBT210}.
	
	{\flushleft \bf GTOT evaluation}.
	From Fig.~\ref{fig:GTOT+RGBT24+RGBT210} (a) we can see that our CAT achieves competitive performance with the state-of-the-art methods.
	In particular, we advance the baseline method RT-MDNet+RGBT with 5.0\%/4.8\% gains in PR/SR. 
	Comparing with the state-of-the-art method MANet, CAT is 0.5\% lower in PR with 5 pixels threshold but 1.0\% higher in PR with 20 pixels threshold.
	Note that our PR score with 20 pixels threshold reaches nearly 100\%, which suggests that almost all frames are tracked by our CAT. 
	As for SR, our CAT is only 0.7\% lower than MANet but about 20 times faster than it.
	It suggests that our CAT achieves a very good balance between tracking accuracy and efficiency.
	Comparing with DAFNet~\cite{gao2019deep}, our PR score with 20 pixels threshold is 1.6\% higher and PR score with 5 pixel is slightly lower but comparable, and SR is 0.5\% higher.
	It should be noted that DAFNet uses complex attention mechanisms for information fusion of two modalities and our CAT does not contain any attention-based fusion.

	{\flushleft \bf RGBT210 evaluation}.
	We also evaluate our CAT on RGBT210 dataset and the results are shown in Fig.~\ref{fig:GTOT+RGBT24+RGBT210} (b). 
	From results we can find that our CAT achieve 0.6\% higher in PR than mfDiMP, which is the winner of VOT2019-RGBT challenge.
	However, our SR is 2.2\% lower than it.
	It is mainly due to two reasons.  
	First, mfDiMP uses the IoU loss to optimize the network, which is beneficial to improving the SR score.
	Second, mfDiMP employs a large-scale synthetic RGBT dataset generated from the training set (9,335 videos with 1,403,359 frames in total) of GOT-10k dataset to train their network, while we only use GTOT dataset (50 videos with 15,000 frames in total) to train our network.
	We will improve the performance of CAT from above considerations in the future.

	{\flushleft \bf RGBT234 evaluation}.
	On RGBT234 dataset, we achieve the best scores in both PR and SR, as shown in Fig.~\ref{fig:GTOT+RGBT24+RGBT210} (c). 
	In particular, our CAT outperforms DAFNet with 0.8\%/1.7\% performance gains, MANet with 2.7\%/2.2\%, and RT-MDNet+RGBT with 7.0\%/7.8\% in PR/SR.
	These results fully demonstrate that the effectiveness of our method.
	

	%
	
	\setlength{\tabcolsep}{4pt}
	\begin{table}[t]
		\begin{center}
			\caption{ PR/SR scores of different variants induced from our method on GTOT, RGBT210 and RGBT234 datasets for verify the effectiveness of the guidance module.}
			\begin{tabular}{ccccccc}
				\hline\noalign{\smallskip}
				&&Baseline&CAT-NS&CAT-NG&CAT-NA&CAT\\
				\noalign{\smallskip}
				\hline
				\noalign{\smallskip}
				\multirow{2}*{GTOT}&PR&0.839&0.876&0.869&0.861&\textbf{0.889}\\
				&SR&0.669&0.708&0.707&0.700&\textbf{0.717}\\
				\hline
				\multirow{2}*{RGBT210}&PR&0.735&0.761&0.760&0.750&\textbf{0.792}\\
				&SR&0.503&0.517&0.516&0.512&\textbf{0.533}\\
				\hline
				\multirow{2}*{RGBT234}&PR&0.734&0.787&0.781&0.773&\textbf{0.804}\\
				&SR&0.483&0.545&0.540&0.535&\textbf{0.561}\\
				\hline
			\end{tabular}
			\label{table:ablation1}
		\end{center}
	\end{table}
	\setlength{\tabcolsep}{1.4pt}
	
	\setlength{\tabcolsep}{4pt}
	\begin{table}[t]
		\begin{center}
			\caption{Compare results of guidance module with FiLM on GTOT and RGBT234 datasets.}
			\begin{tabular}{ccccc}
				\hline\noalign{\smallskip}
				&&Baseline&CAT-FiLM&CAT\\
				\noalign{\smallskip}
				\hline
				\noalign{\smallskip}
				\multirow{2}*{GTOT}&PR&0.839&0.848&\textbf{0.889}\\
				&SR&0.669&0.703&\textbf{0.717}\\
				\hline
				\multirow{2}*{RGBT234}&PR&0.734&0.783&\textbf{0.804}\\
				&SR&0.483&0.540&\textbf{0.561}\\
				\hline
			\end{tabular}
			\label{table:ablation2}
		\end{center}
	\end{table}
	\setlength{\tabcolsep}{1.4pt}
	
	\setlength{\tabcolsep}{4pt}
	\begin{table}[t]
		\begin{center}
			\caption{PR/SR scores of different variants induced from our method on GTOT dataset for verify the effectiveness of hierarchical design. \checkmark means adding parallel challenge-aware branches in this layer.}
			\begin{tabular}{cccccc}
				\hline
				&conv1&conv2&conv3&PR&SR\\
				\hline
				Baseline &&&& 0.839& 0.669\\
				CAT-v1 &\checkmark & & & 0.877&0.701\\
				CAT-v2 & &\checkmark & & 0.878&0.712\\
				CAT-v3 & & & \checkmark&0.880 &0.706\\
				CAT-v4 &\checkmark &\checkmark & &0.865 &0.708\\
				CAT-v5 &\checkmark & &\checkmark &0.877 &0.710\\
				CAT-v6 & & \checkmark &\checkmark  &0.876 &0.704\\
				\hline
				CAT &\checkmark &\checkmark  &\checkmark  &\textbf{0.889} &\textbf{0.717}\\
				\hline
			\end{tabular}
			\label{table:ablation3}
		\end{center}
	\end{table}
	\setlength{\tabcolsep}{1.4pt}

	\subsection{In-depth Analysis of The Proposed CAT}
	\subsubsection{Ablation study}
	To verify the effectiveness of main components of the proposed tracker, we carry out the ablation study on the GTOT, RGBT210 and RGBT234 datasets. 
	We implement three special versions of our method to verify the role of the guidance modules, they are: 1) CAT-NS, that removes the gate operations in all guidance modules; 2) CAT-NG, that removes the gate operations and convolutional operations in all guidance modules, in other words, that adds the features of another modality directly. and 3) CAT-NA, that removes all guidance modules.
	Table~\ref{table:ablation1} presents the comparison results, 
	we can find that all the ablation experiments perform better than the baseline, which fully demonstrates the effectiveness of the proposed method,  
	and CAT-NS is lower than CAT which proves that the gate mechanism can effectively suppress the propagation of noisy information, 
	which can also be verified from the visualization of the thermal crossover (TC) in Fig.~\ref{fig:motivation}.
	Comparing CAT-NS with CAT-NG, we can find that CAT-NS is better than CAT-NG , and all of them is higher than CAT-NA in all datasets, which shows the effectiveness of introduce information from another modality in modality-specific branches.
	%
	
	We also compare guidance module with FiLM on GTOT and RGBT234 datasets to verify the effectiveness of proposed method, which can be seen in Table~\ref{table:ablation2}, our guidance module achieves superior performance over FiLM on GTOT and RGBT234.
	
	To verify the effectiveness of our hierarchical design, we add the parallel challenge-aware branches in one, two or all of convlutional layers in the backbone network, and the evaluation results are shown in Table~\ref{table:ablation3}.
	We can find that no matter how parallel branches are added, the performance is significantly improved compared to the baseline, further demonstrating the effectiveness of the challenge-aware module.
	When adding the challenge-aware branches in all three layers at the same time, we can obtain the best performance, which fully demonstrates the effectiveness of the hierarchical challenge-aware target representations.

	\subsubsection{Runtime analysis}
	Our tracker is implemented in pytorch 0.4.0, python 2.7, and runs on a computer with an Intel Xeon E5-2620 v4 CPU and a GeForce RTX 2080Ti GPU card.
	%
	%
	Our tracker achieves about 20 FPS, which is slower than the baseline RT-MDNet+RGBT (nearly 30 FPS), but we achieve much better tracking performance on three RGBT tracking benchmark datasets.
	
	\section{Conclusion}
	In this paper, we have proposed a novel deep framework to learn target appearance representations under different challenges even in the case of insufficient training data for RGBT tracking.
	We propose to use parallel and hierarchical challenge-aware branches to represent target appearance changes under certain challenges while maintaining a low computational complexity.
	Extensive experiments on three benchmark datasets demonstrate the effectiveness and efficiency of the proposed method against the state-of-the-art trackers.
	In the future, we will explore more challenges to enhance target representations and study more suitable structures to handle different kinds of challenges to improve the effectiveness and efficiency.
	
	%
	%
	\bibliographystyle{splncs04}
	\bibliography{egbib}
\end{document}